# You Are in Control of Your State: Why Human Outcomes Are Controllable Through Causal State Intervention

**Suraj Biswas · Saurav Gupta · Pritam Mukherjee**

*ORCID: Suraj Biswas - https://orcid.org/0009-0008-1727-8179; Pritam Mukherjee - https://orcid.org/0009-0007-9018-4083*

*Independent Researchers in Behavioural Modelling, Causal Inference, and Genomics*

*Assessli · Dots-In*

**Abstract.** A central puzzle for the behavioural sciences and for human-facing artificial intelligence is the persistence of within-person variability. The same individual, presented with the same observable input, produces different outcomes on different occasions. Different individuals presented with the same input produce divergent outcomes that no observable covariate fully predicts. This paper develops a position that places that variability where it belongs, in the dynamic latent state of the person, and argues that human outcomes are controllable in a precise and operational sense, but only through interventions that target the state and its weighting at the moment a decision is being formed.

We define a state as the time-indexed weighting vector over the dimensions that govern how an individual's biology, physiology, and neuropsychology process the next event into a decision and an outcome. We argue that the relationship between state, decision, and outcome is causal rather than correlational. The weighting vector is dynamic at sub-daily timescales. The conscious channel through which outcomes are reportable is a narrow attentional bottleneck whose contents are themselves state-dependent. Taken together, these claims imply that the outcome of a given event is neither fully predictable from observable covariates nor uniformly controllable. It is controllable, conditionally, on the state-trajectory at the time of intervention.

We motivate the framework with six lines of established evidence: causal hierarchies in behavioural inference, allostatic and predictive-coding accounts of state, the bandwidth gap between sensory input and reportable cognition, persistent within-person variability in computational psychiatry, chronobiological structure of within-day variation, and the failure modes of correlational personalisation systems documented in 2020 to 2026. We describe a 24-month observational base from a deployed behavioural platform that spans more than 200,000 consented users across four occupational personas, with research carried out from 2023 to 2026 by the authors and their team. At the aggregate qualitative level, the observed patterns are consistent with the framework. We derive seven testable predictions, list six operational requirements for state-aware systems, and discuss implications for digital health, education, AI personalisation, and personal agency. The paper is a position contribution, not a clinical or pre-registered empirical study. The empirical observations are motivating context and are reported at the aggregate level only.

# 1 Introduction

## 1.1 Three short scenes

A student receives the same examiner's comment on two essays a week apart, in identical wording. The first time she rewrites for two hours and submits a stronger draft. The second time she closes the document, sleeps badly, and does not return to it for three days. The event class is the same. The person is the same. The outcome is not. A founder sits through two pitch sessions in the same month and gets the same kind of pushback from investors. The first session ends with a strategic reframing on the way home. The second ends with three days of avoidance and a missed follow-up email. A developer reads the same code-review comment on two pull requests, written by the same colleague in nearly the same wording. The first time, the comment lands as a useful steer. The second time, it lands as a personal slight. These are not exotic examples. They are the texture of ordinary cognitive life, and any honest theory of human outcomes has to begin from them.

The within-person variability literature is large and old, and the puzzle is recognised across psychology, behavioural economics, and computational psychiatry [7, 24, 12]. What remains contested is what we should locate the variability in, and what to do about it. A familiar move is to add covariates: measure mood, measure sleep, measure time of day. The residual variability shrinks a little. A second familiar move is to add interactions: posit that feedback effects depend on personality, baseline stress, and social context. Interactions help a little more. Even then, the residual within-person variability is stubborn. In data drawn from real lives, with dense predictors and careful modelling, the same person under nominally similar conditions does different things on different occasions. This paper argues that the residual is not noise around a mean. It is the signature of a real and characterisable feature of human cognition, and the right object to put at the centre of an account of human outcomes is the latent state of the person at the moment the input arrives.

## 1.2 What this paper claims

We make three claims, in tightening order of strength. The first is descriptive. Human outcomes are state-conditional. The function that maps an observable event class to an outcome class is not a function of the event alone but of the event evaluated against a time-indexed weighting vector that constitutes the person's state. The second claim is mechanistic. The relationship between state, decision, and outcome is causal, not correlational, in the sense made precise by Pearl [28] and Bareinboim et al. [1]. Observational distributions over event-outcome pairs underdetermine the interventional distribution that any controller, human or artificial, would need to predict the effect of a change. The third claim is prescriptive. This picture supports a meaningful sense in which people can control their outcomes. Not by willing the outcome directly. Not by adding effort. By intervening on the state and on the timing of intervention. Control of this kind is real, bounded, and operational. It has measurable signatures, and it can in principle be supported by intelligent systems that are built for it.

## 1.3 Where this paper sits

The present paper is a companion to Biswas [2], which argued that any system claiming to model an individual human must instantiate four jointly necessary properties: persistent state representation, explicit causal structure, episodic temporal memory, and welfare-aligned objectives. That earlier paper made the case at the level of system specification. The present paper makes the case at the level of the person. If the causal architecture paper described what a model of a human must contain, the present paper describes what is in fact the case about the human on which any such model would have to operate. We will recur to four of the same primitives. We will not assume the system specification.

## 1.4 What this paper does not claim

We do not claim that any specific intervention works. We do not claim that the deployed platform from which we draw motivating observations is a clinical instrument. We do not report clinical or neuropsychological measurement. We do not report data from validated instruments. We do not provide pre-registered hypothesis tests. The empirical observations we cite are aggregate, qualitative, and motivating. They are reported to ground the framework in a real deployment context, not as findings in the technical sense. A separate empirical companion paper, when the analysis pipeline is complete and the disclosure boundary settled, will report quantitative findings under the appropriate methodological controls.

### 1.5 Roadmap

Section 2 develops the formal notion of state and describes what is meant by state weighting. Section 3 places the framework against six strands of relevant literature. Section 4 develops the bandwidth-and-bias argument. The conscious channel through which a person reports a decision is narrow, state-dependent, and a poor place to look for the actual driver of an outcome. Section 5 describes the observational base and the kinds of patterns that emerge from it. Section 6 lists six operational requirements for any state-aware system. Section 7 derives seven falsifiable predictions. Section 8 discusses implications and limitations. Section 9 outlines future work. Section 10 concludes.

## 2 State as a Dynamic Weighting Variable

### 2.1 A working definition

By state we mean the time-indexed weighting vector that determines how the next event is evaluated, how the biological and physiological substrate responds, and how a decision is produced. The definition is formal:

**Definition (State).** For an individual I at time t, the state S_t is the latent vector that, jointly with the event X_t and the individual's prior history H_{t-1}, determines the conditional distribution of the next decision A_t and outcome Y_t. State has three properties. It is dynamic: S_t evolves through specified update operators applied to intervening events and biological processes. It is multi-dimensional: its dimensions span biological, physiological, neuropsychological, and contextual factors. It is weighting: each dimension contributes to the evaluation of X_t with a coefficient that itself depends on S_t.

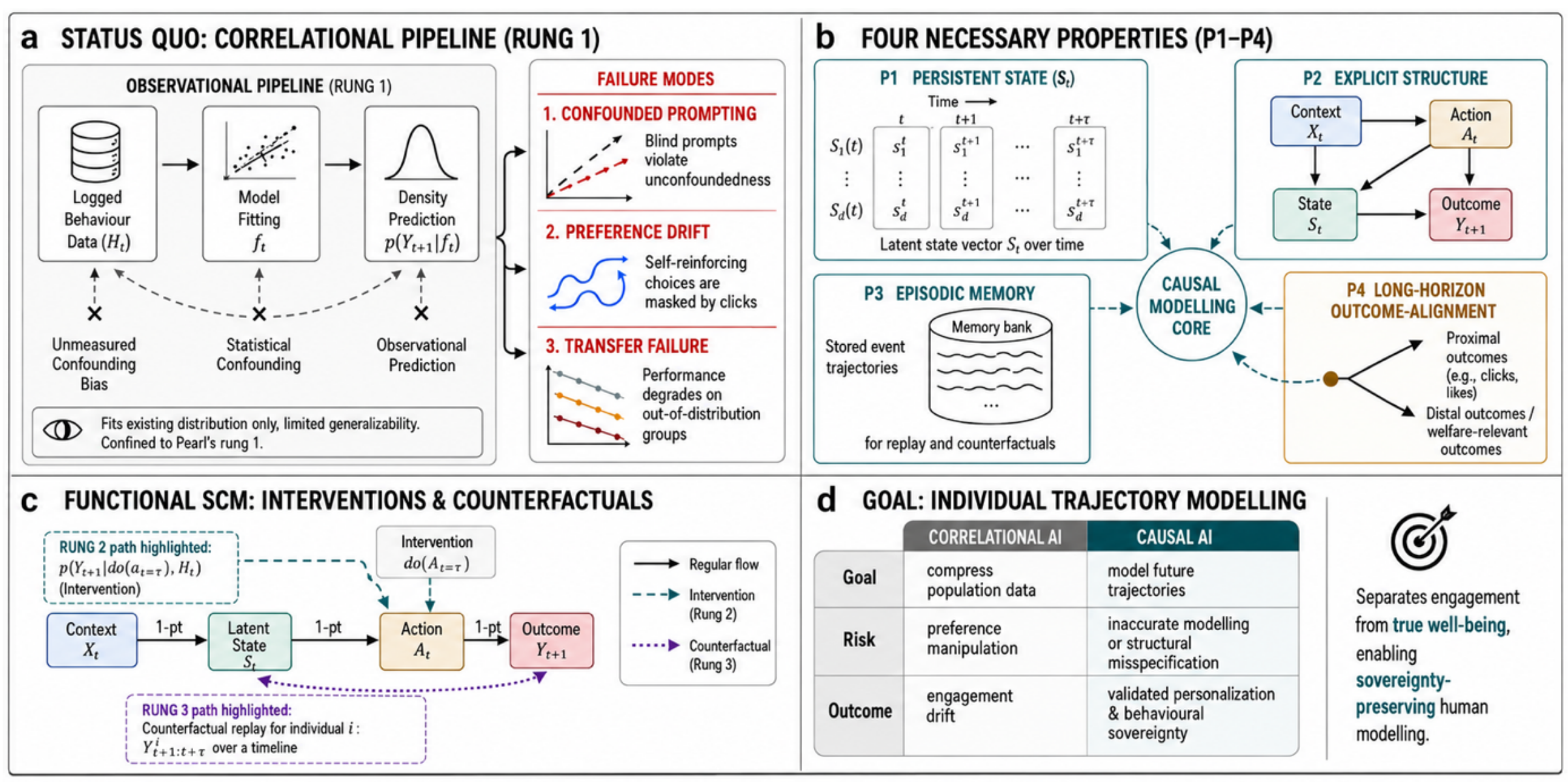

**Figure 1.** *Same observable input, different states, different outcomes. The event is held constant. The weighting vector S differs across three illustrative configurations. The outcome class diverges accordingly. The arrow from input to state is conditional rather than strictly causal because the state pre-exists the input. The arrow from state plus input to outcome is the causal claim.*

### 2.2 Why a weighting variable, not a feature vector

A common move in machine learning is to represent the person as a feature vector and to predict outcomes as a function of features and context. That move treats the person as a constant and the context as a variable. Our proposal is the opposite. The person, regarded as a feature vector, is not constant. The dimensions that matter for any specific event are themselves weighted by the state at the moment of evaluation. A student with the same conscientiousness score and the same prior performance is differently responsive to identical feedback on different mornings. The difference is not noise around a mean. It is the action of a real time-varying weight on the way the input gets processed.

Three traditions in the literature come close to this picture without quite stating it. Predictive processing and active inference [9, 30] model the brain as a system that holds and updates a generative model of its world, with the current state of that model determining what counts as a prediction error and what response is selected. Allostasis [36, 23] frames biological stability as achieved through change rather than through holding parameters constant. Computational psychiatry [13, 10] phenotypes individuals through generative-model parameters rather than through clustering labels. The framework here is in all three lineages. We are sharper, however, about three things. The state is multi-channel. It has biological, physiological, and neuropsychological components, each of which can in principle be intervened on. The weighting is causal. The same event under different weightings produces different downstream biology and physiology, not merely different reported affect. The result is operational. State can be tracked, and the controllability of outcomes is conditional on the state-trajectory.

### 2.3 The causal path

Figure 2 depicts the causal path. State S_t propagates through three intermediate channels. Biology covers HPA-axis dynamics, autonomic regulation, and endocrine signalling. Physiology covers heart-rate variability, sleep architecture, activity, and energy availability. Neuropsychology covers attention allocation, appraisal, and working memory. The three channels jointly shape the decision A_t, which produces an outcome Y_t. The outcome feeds back, with a delay, into the next-period state S_{t+1}. Each of the three intermediate channels is in principle an intervention point. The decision itself is also an intervention point. As we argue in Section 4, it is the least efficient of the four, because the conscious channel through which the decision is produced is narrow, biased, and itself state-determined.

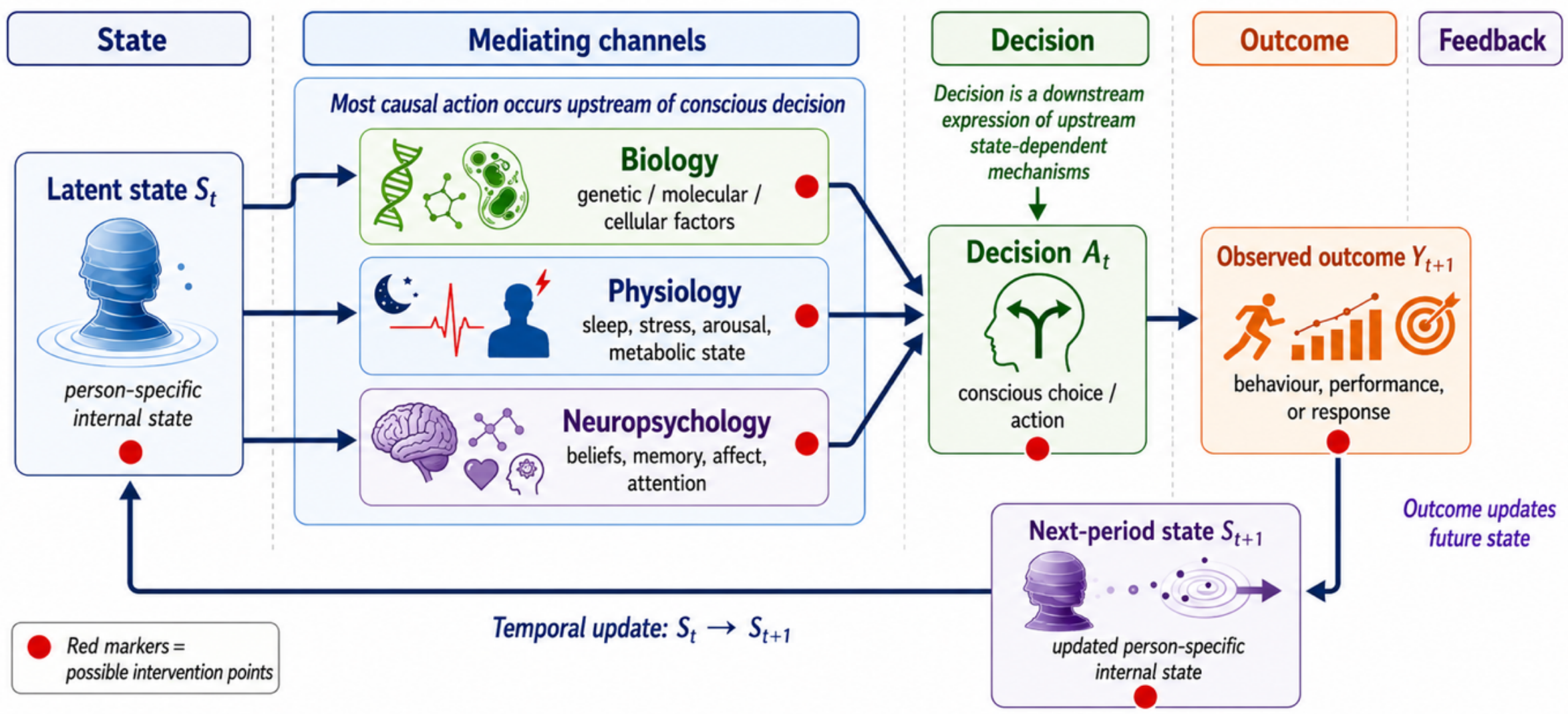

**Figure 2.** *The causal path from state to outcome, with feedback. The state propagates through three channels (biology, physiology, neuropsychology) to the decision, and through the decision to the observed outcome. The outcome updates the next-period state. Red markers indicate intervention points. The bulk of the causal action happens upstream of the conscious decision.*

### 2.4 State has structure, not just position

A useful corollary follows. Because state is a weighting vector and not a position in feature space, the same person can have the same observable features at two different times and still be in different states. The reason is that the weighting on those features is different. This is sometimes called the equifinality problem in developmental and personality psychology: distinct underlying configurations can produce indistinguishable surface profiles [5]. It is the source of much of the noise that haunts correlational personalisation systems. It is the reason we have to put the latent weighting, not the observable profile, at the centre of any account of why outcomes diverge.

### 2.5 State-update dynamics

What does it mean for S_t to evolve? Three kinds of update operate together. Slow updates come from circadian and ultradian rhythms, sleep architecture, hormonal cycles, and seasonal variation. These set the baseline structure of within-day variation. Medium-speed updates come from social interactions, work events, exercise, food, and exposure to media. These ride on top of the baseline and shift the state by perceptible amounts within minutes to hours. Fast updates come from the immediate processing of new events. A piece of feedback, a sudden noise, a thought about an old commitment can shift the state in seconds. The combined trajectory is not random. It has structure that can be characterised at the individual level, even when group-level summaries miss it.

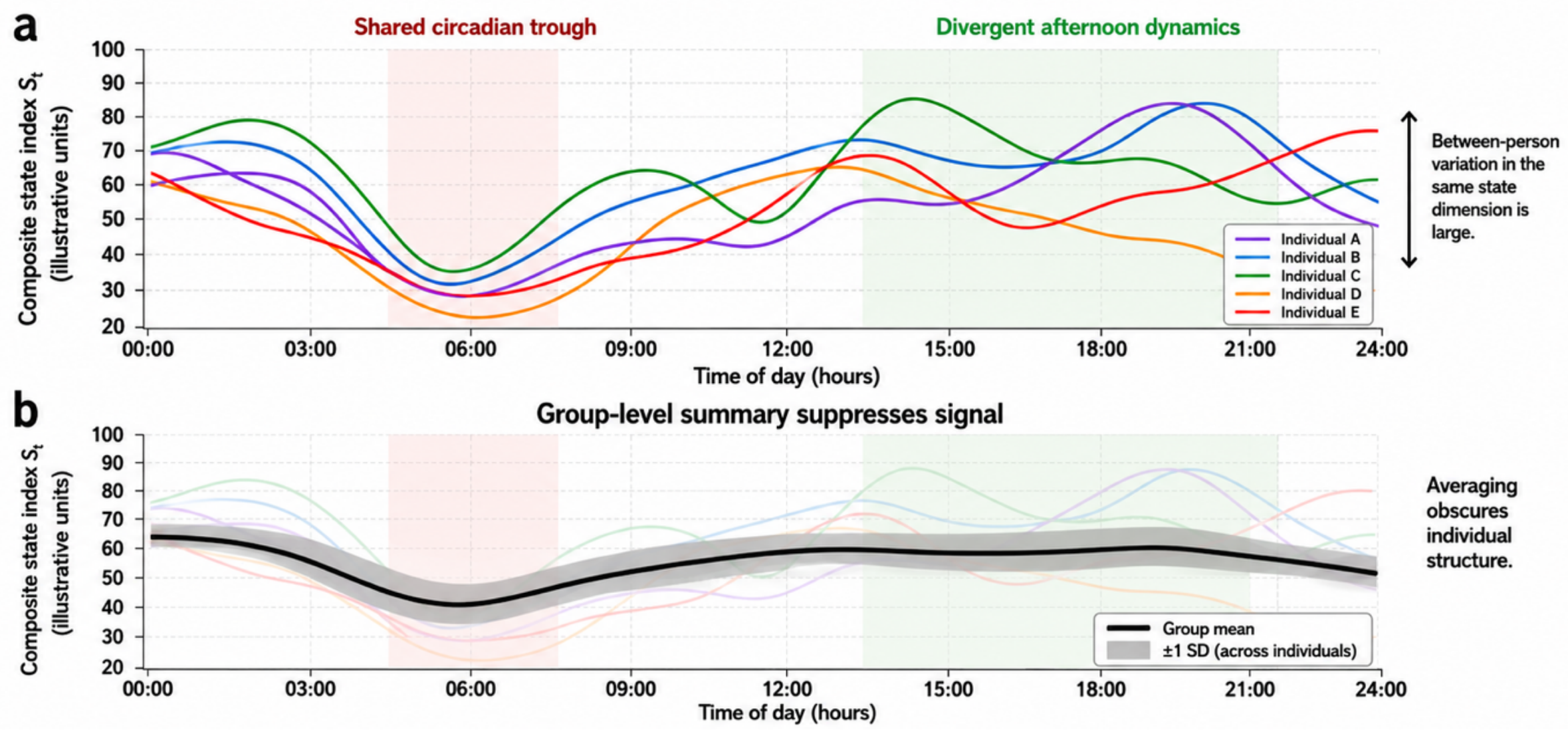

**Figure 3.** *Within-day state oscillation across five illustrative individuals (schematic). The same composite state index varies through the day in patterns that share some structure (the early-morning trough) and diverge in others (afternoon dynamics). Between-person variation in the same dimension is large enough that group-level summaries discard most of the signal.*

### 2.6 State has correlation structure that differs across people

A further property of state, often missed by treatments that focus on single dimensions, is that the correlation structure among dimensions itself varies across individuals and across groups. Figure 4 shows what this looks like at the level of an illustrative four-persona comparison. Arousal correlates with cognitive load in some patterns and not in others. Sleep debt couples to confidence in some groups more than others. The pattern matters because intervention design depends on it. An intervention that improves sleep is also acting on confidence and goal clarity in some people more than in others, through the within-person correlations among dimensions. This is one of the reasons that intervention effectiveness is so heterogeneous in the behaviour-change literature even when the protocol is held fixed [32, 25].

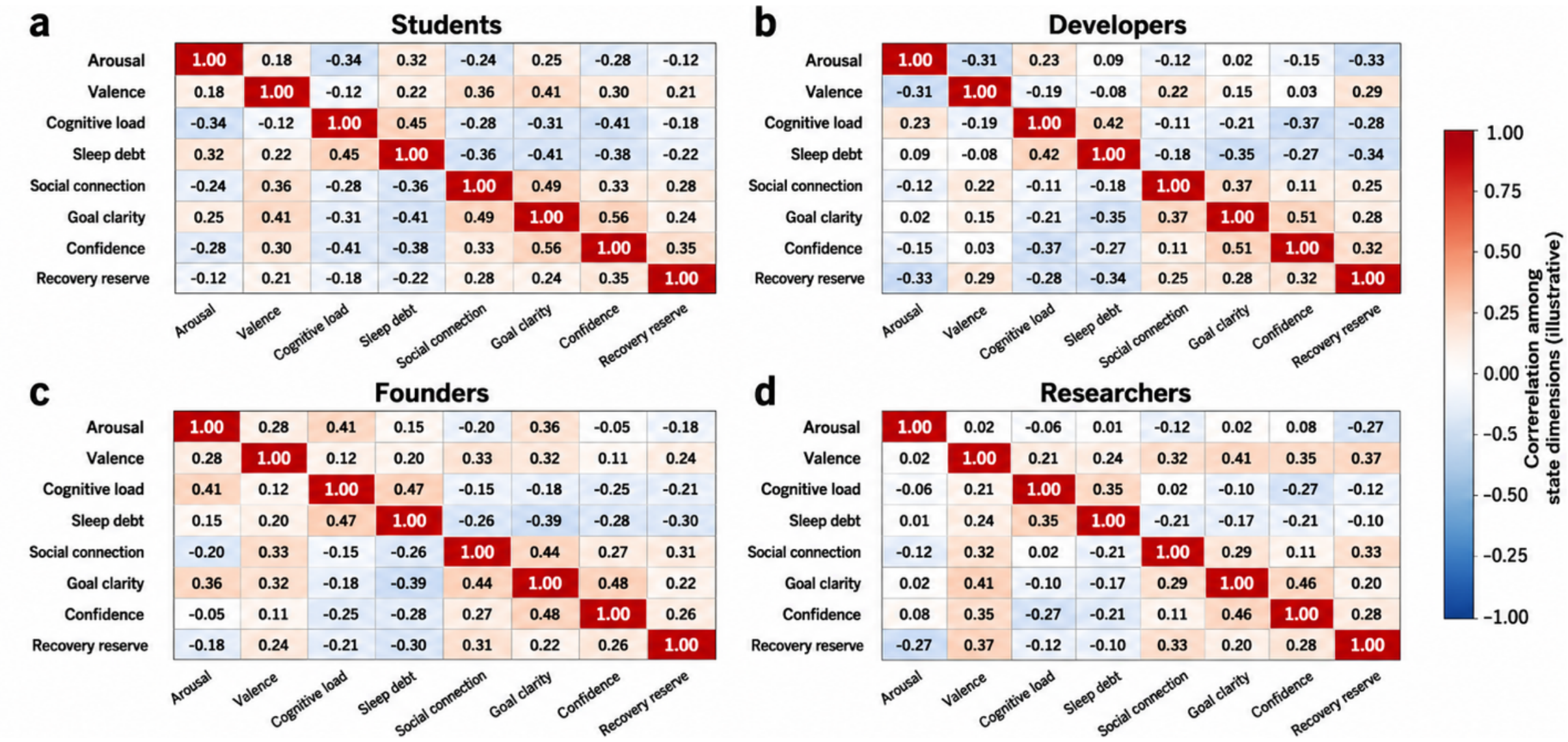

**Figure 4.** *Illustrative state-dimension correlation structure across the four personas. The same dimensions are weighted differently in different groups, and the correlations among them carry information that a group average discards. The matrices are illustrative, not measurements.*

## 3 Background and Related Work

We organise the literature into six strands. We are deliberately brief on each, because the relevant 2020 to 2026 syntheses are surveyed in detail in Biswas [2], and we cross-reference rather than rehearse.

### 3.1 Causal inference and the limits of correlation

Pearl [28] and Bareinboim et al. [1] formalised the three-rung causal hierarchy. Associations, interventions, and counterfactuals are not mutually reducible. Inferences at higher rungs require structural information beyond what observational distributions can supply. Schölkopf et al. [35] extended the picture to machine learning, arguing that representations that support transfer and intervention require explicit causal structure. Richens and Everitt [31] sharpened the point. Any agent that generalises under distributional shift must have implicitly acquired an approximate causal world model. For the present paper, the relevant consequence is that an observational distribution over event-outcome pairs underdetermines the effect of changing the state. We cannot read off, from a corpus of observed behaviours, the function from intervention to outcome. We have to model it.

### 3.2 Predictive processing and active inference

Friston [9, 10] and Pezzulo et al. [30] provide the generative-model account of cognition we lean on most heavily. In that picture, the brain is continuously generating predictions about its sensory input, evaluating prediction errors, and updating either its model or the world to minimise free energy. The state, in our sense, corresponds roughly to the parameters of the current generative model. The weighting is what predictive processing calls precision: the confidence with which predictions are held. High-precision priors close down the perceptual channel. Low-precision priors open it. We adopt the structural commitments of this tradition (causality, intervention, generative modelling) and remain agnostic about the strongest free-energy interpretations.

### 3.3 Allostasis and the biological state

Allostasis [36, 23] is the principle that biological systems achieve stability through change rather than through holding parameters constant. The state, in our sense, includes the allostatic load (the running cost of repeated adaptive responses) and the biological reserves available to mount the next response. Recent work has connected allostatic markers (HRV, cortisol slope, metabolic substrates) to behavioural readiness for cognitive challenges [11, 16] and to the differential response of individuals to identical stressors [14]. The framework here treats the allostatic state as one of the three intermediate channels through which the state propagates to outcomes.

### 3.4 The attentional bottleneck

A well-established result in cognitive science is that the conscious channel through which decisions are reported is narrow. The brain receives an estimated $10^{11}$ neural events per second across all sensory modalities [17], but conscious processing is limited to a serial stream estimated at 40 to 60 bits per second [6, 19, 20]. What passes through the bottleneck is not a representative sample of the input. It is heavily filtered by current goals, current affect, and current state, in ways that have been documented in the attention and working-memory literature for decades. Section 4 develops the consequence: that the verbal narrative a person reports about why they made a decision is a poor guide to the actual causes, and that intervention on the verbal narrative is correspondingly inefficient.

### 3.5 Chronobiology and within-day structure

A literature that does not always communicate with the cognitive-architecture literature is chronobiology. The within-day structure of cognitive performance, mood, and decision behaviour has been mapped in detail. Roenneberg et al. [34] documented stable inter-individual differences in chronotype that predict daily peaks and troughs of performance. May and Hasher [22] showed that the time of day at which a cognitive task is performed interacts with chronotype to determine accuracy on attention-demanding tasks. Folkard and Akerstedt [8] characterised the joint effect of circadian phase and homeostatic sleep pressure on sustained performance. The relevance to our framework is direct. The trajectory of the state through a day is not free-form. It has a structure that is partly innate, partly entrained by behaviour and environment, and partly shifted by recent history. Any state-aware system has to respect this structure. Most personalisation systems do not.

### 3.6 Computational psychiatry and within-person variability

Computational psychiatry has explicitly used generative-model parameters to phenotype individuals [13, 10]. Recent reviews [21, 4] document that data-driven phenotyping produces classifications that do not transfer across cohorts, while theory-driven phenotyping (recovery of generative-model parameters) does. The framework here is in the latter tradition. State, as we use the term, is a generative-model parameter, not a clustering label.

### 3.7 The gap

These strands collectively describe components of the picture but rarely combine them at the level of operational claim. Predictive processing has the right structural commitments but does not always translate into deployment-relevant interventions. Allostasis has the right biological grounding but is rarely connected to behavioural prediction at the level of the individual day. Causal inference has the right mathematical apparatus but is typically applied at the population level. Cognitive science has the bottleneck argument but does not generally connect it to the design of interventions. Chronobiology has the within-day structure but is rarely fused with cognitive-architecture commitments. Our contribution is to combine the strands at the level of operational claim about individual outcomes.

## 4 Bandwidth, Bias, and Why the Conscious Channel Is the Wrong Lever

### 4.1 The bandwidth gap

A standard estimate of the input to the human nervous system places it on the order of $10^{11}$ events per second across all sensory modalities, counting neural firings [17]. The output of conscious cognition, measured in bits per second of reportable information, is on the order of 40 to 60 [6, 19, 20]. The gap is not seven, or nine. It is roughly nine orders of magnitude. Figure 5 depicts the funnel. What comes out of the bottleneck is not chosen by the conscious agent in any sense that would be useful to an interventionist. It is selected by attentional and pre-attentive filters whose settings are determined by the current state.

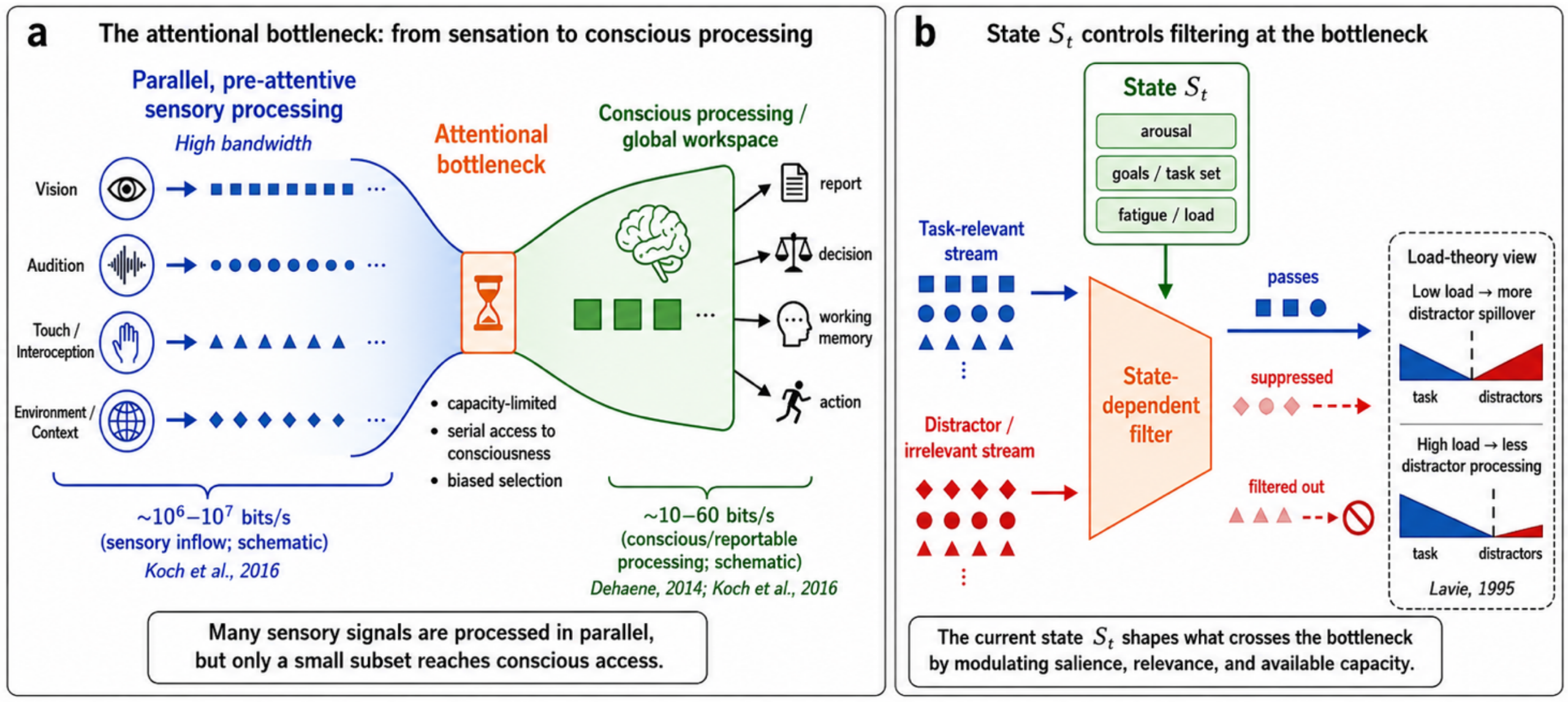

**Figure 5.** The attentional bottleneck and state-dependent filtering. Sensory input is parallel and pre-attentive. Conscious processing is serial, capacity-limited, and biased. The state $S_t$ controls what passes the bottleneck. Bandwidth estimates from Koch et al. [17], Dehaene [6], Lavie [19].

### 4.2 What this implies for the verbal narrative

A person who says, after a decision, "I closed the document because I was tired" is reporting a verbal narrative that has itself passed through the bottleneck. The narrative is not necessarily inaccurate, but it is necessarily compressed, and the compression is state-dependent. A standard finding in the introspection literature is that self-reports about the causes of behaviour are reliably wrong in predictable ways [26, 37]. The reasons a person gives are a function of the reasons that survived the bottleneck, not of the reasons that actually drove the decision. For an interventionist, this is consequential. If we ask a person to change a behaviour by reasoning with them about why they do it, we are intervening on the most compressed and most state-dependent layer of the causal chain. It is usually the least efficient layer to intervene on, even when the person agrees with us about the diagnosis.

### 4.3 A worked example: perceptual load

A concrete illustration comes from the perceptual-load literature. Lavie [19, 20] showed that the amount of distractor processing that occurs while a person is performing a primary task depends on the perceptual load of the primary task. Under low load, irrelevant distractors are processed and can capture attention. Under high load, they are filtered before they reach conscious processing. The current state of the perceiver, including arousal and goal precision, modulates the threshold at which filtering kicks in. Two implications follow. The same distractor has different effects in different states. And the person's introspective report about why they were distracted is generated post hoc, with little access to the filtering mechanism itself.

### 4.4 Motivated cognition and the trouble with deliberation

A further consequence of the bandwidth gap is well-documented in the social psychology of motivated cognition. Kunda [18] showed that people arrive at desired conclusions through directional reasoning while maintaining the subjective sense that they are reasoning impartially. The state-weighting framework reads this as a special case. When the state weights a particular outcome class as more desirable, the limited bandwidth of the conscious channel is preferentially allocated to evidence that supports that outcome, and the resulting verbal narrative looks like deliberation but is more accurately described as state-conditioned filtering. The practical implication is that deliberation about a decision performed in an unfavourable state is not a corrective. It is a recapitulation of the state's preferred answer in the form of a reasoned argument. The corrective is intervention on the state, not more deliberation.

### 4.5 Where the leverage actually is

Figure 2, revisited, makes the implication concrete. The four intervention points in the causal chain are not equally efficient. Interventions on the biological channel (sleep, circadian rhythm, exercise, nutrition) and on the physiological channel (HRV-mediated regulation, breath, posture) have effects that propagate forward to attention, appraisal, and decision. Interventions on the neuropsychological channel (attention training, reframing, exposure) are slower-acting but well-established. Interventions on the decision itself (telling the person to choose differently) are the least efficient, because they require the conscious channel to carry information against the gradient of the current state-weighting. None of this is news to clinical psychology. It is news to a great deal of digital personalisation.

### 4.6 A note on the 10-bit claim

The popular formulation that the conscious channel processes "ten bits while receiving billions" is a paraphrase of an extrapolation in Norretranders [27] and is widely repeated but not directly empirical. The peer-reviewed numbers we cite are firmer. The upper bound on conscious throughput is in the tens of bits per second in attention experiments [19, 20]. The neural-firing rate estimate is in Koch et al. [17]. The qualitative point survives. The gap is enormous, and what passes is state-dependent.

## 5 Observational Base

### 5.1 What the deployment is and is not

The framework above was developed alongside, and informed by, a deployment of a behavioural platform with consenting users in India. The research period covers 2023 to 2026. The observational window from which we draw motivating evidence is the most recent 24 months. The deployment is a product, not a clinical instrument. Users sign up to receive personalised behavioural support. They consent at registration to the use of their de-identified data for research and product development. They receive personalised content, scheduled prompts, and reflective exercises. We summarise the cohort and the data sources in Figure 6 and report observations at the aggregate qualitative level only. We do not report quantitative effect sizes, pre-registered hypotheses, or clinical measurements. We report what the deployment looked like, why it informed the framework, and what features of the data are consistent with the framework's predictions.

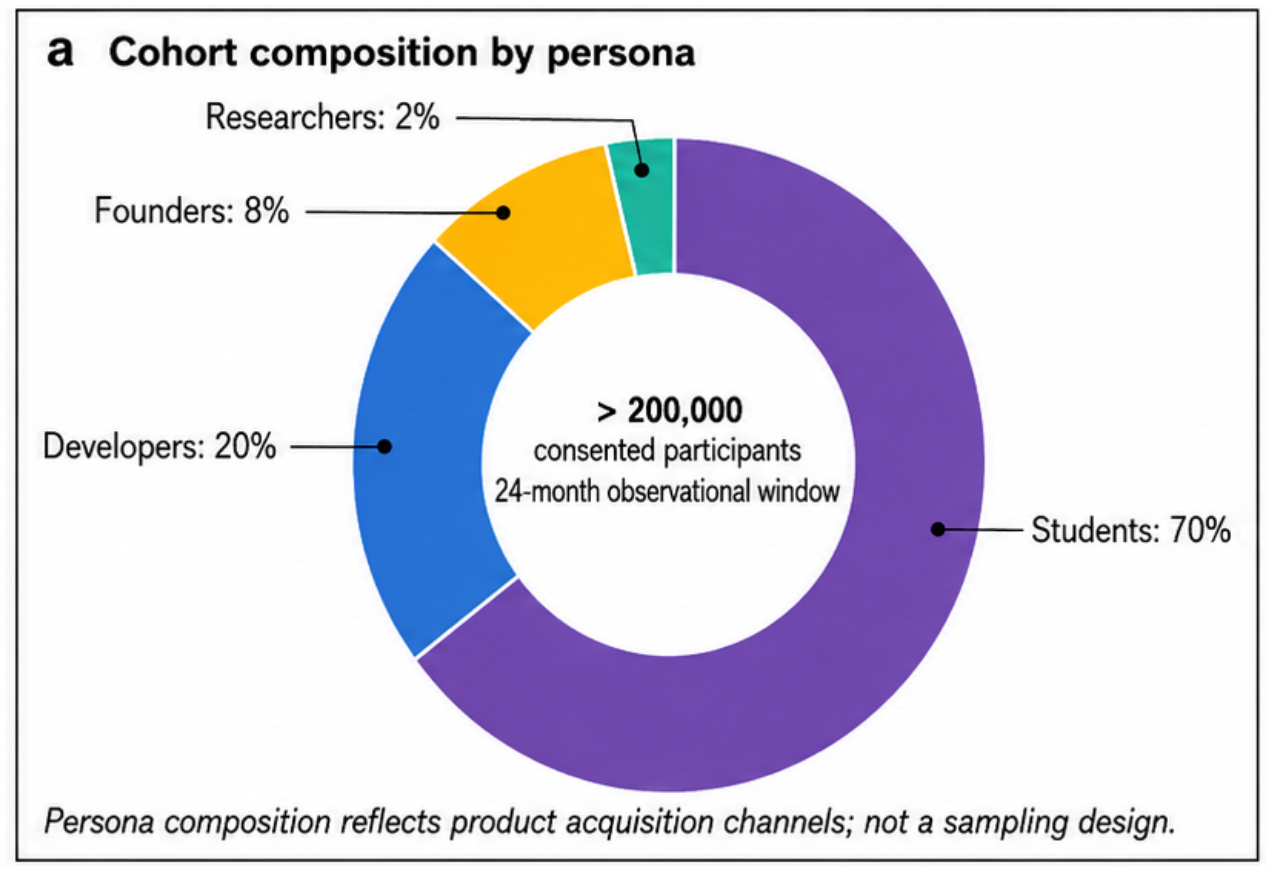

**b Cohort and study metadata**

| Item | Description |
|---|---|
| Sample base (consented) | > 200,000 participants |
| Observational window | 24 months |
| Research period | 2023–2026 |
| Persona strata | Students (70%), Developers (20%), Founders (8%), Researchers (2%) |
| Study design | Observational, product-embedded deployment |
| Data sources | Platform telemetry; self-report; scheduled and on-demand responses |
| Inference status | Aggregate qualitative findings reported here |
| Quantitative analysis | Deferred to separate empirical companion |
| Role of persona | Observational stratifier, not a randomised condition |
| Key caveat | Cohort heavily skewed toward students; cross-persona inferences should be bounded accordingly |

**Figure 6.** *Observational base supporting the framework. (a) Cohort composition by persona across more than 200,000 consented participants observed over 24 months. (b) Cohort and study metadata. The deployment is observational and product-embedded. Aggregate qualitative findings are reported here. Quantitative analysis is deferred to a separate empirical companion.*

### 5.2 Cohort stratification

The cohort is stratified across four occupational personas. Students account for approximately 70 per cent of the base, developers approximately 20 per cent, founders approximately 8 per cent, and researchers approximately 2 per cent. The stratification is a product of the platform's acquisition channels and is not the result of a sampling design. The composition is heavily skewed toward students, and the inferential weight we place on any cross-persona comparison must be bounded accordingly. We treat the persona variable as an observational stratifier rather than as a randomised condition.

### 5.3 What kinds of data the platform produces

Two streams of data flow from the deployment. The first is platform telemetry: time-stamped logs of user actions, content interactions, prompt responses, and within-platform navigation. The second is self-report: scheduled and on-demand reports of mood, decision quality, satisfaction, life events, and qualitative reflections on intervention exposure. Neither stream is a clinical measurement. We do not collect heart-rate variability, cortisol, sleep architecture, or scores on validated neuropsychological instruments. We collect the behavioural and subjective traces a user leaves while using a digital product, and we collect them at the cadence the product affords. A common figure quoted internally for total event counts is on the order of tens of thousands of per-user events per active day, summed across telemetry types. This figure is platform-scale and not a measurement claim about cognitive processing.

### 5.4 Outcome divergence across personas

Figure 7 depicts, schematically, the kind of outcome-divergence pattern that is consistent with the framework. For a generic class of high-stakes feedback events, the distribution over outcome classes (withdrawal, rumination, reframing, action, integration) differs across the four personas. The pattern in Figure 7 is illustrative and not a measurement. It is shown to anchor the kind of finding the framework predicts. The deeper pattern, which is consistent with the framework, is that within each persona the variance is large, and that an event class plus a persona is insufficient to predict the outcome class. What is required is the state-trajectory at the moment the event arrives. This is the framework's central observational claim.

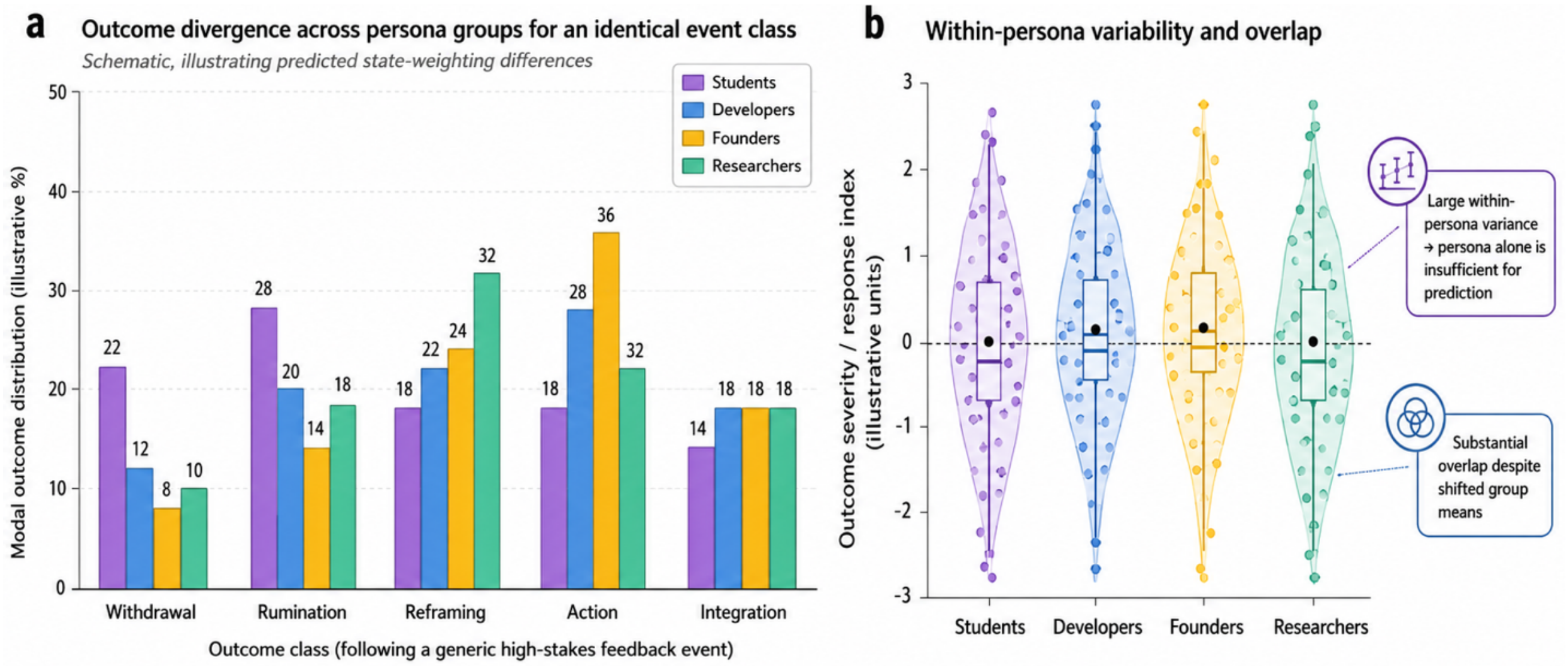

**Figure 7.** *Outcome divergence across persona groups for an identical event class. The bars are illustrative percentages, not measurements. They are presented to anchor the kind of finding the framework predicts: that the same event class produces materially different outcome distributions across personas, and that within each persona the variance is large enough that the persona alone is insufficient to predict the outcome.*

### 5.5 Intervention exposure and the timing question

Users receive personalised cues from the platform. The platform decides what to surface and when. We will not describe the architecture by which those decisions are made. The architecture is the subject of a separate set of patent applications. We will describe the behavioural question the platform raises. When a cue is delivered, the user is in a state. The state at the moment of delivery is, in our framework, a critical determinant of whether the cue produces the intended behavioural change. A useful intervention delivered at an unfavourable moment is approximately useless. A modest intervention delivered at a state-transition window can have an outsized effect. Figure 8 depicts the schematic logic. State trajectories pass through windows of high and low controllability. Well-timed intervention is the controllability-relevant variable, not intervention amplitude.

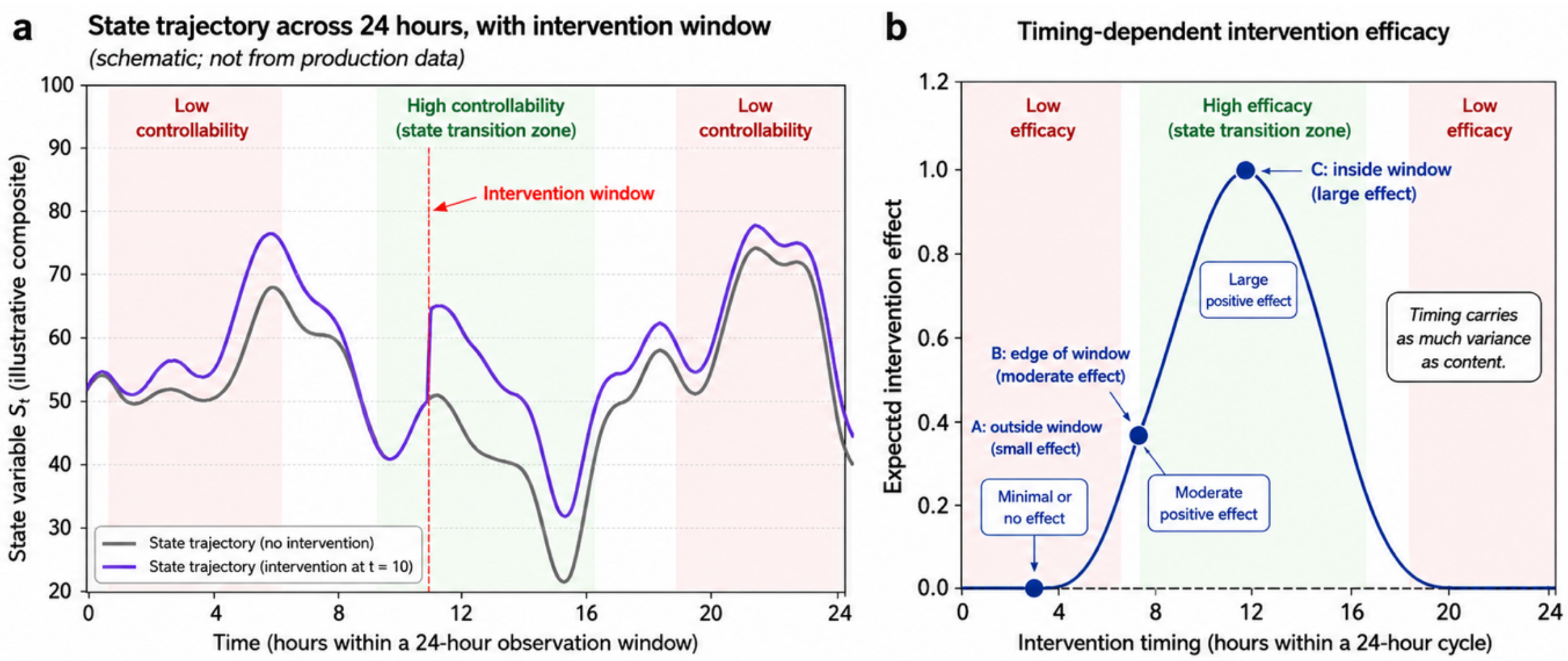

**Figure 8.** *State trajectory across 24 hours, with intervention window. The trajectory is a schematic composite, not from production data. The point is structural. Not all moments are equally amenable to intervention. State-transition zones are windows in which a modest intervention can shift the trajectory. Outside those windows, the same intervention has small or no effect. The implication is that timing carries as much of the variance as content.*

### 5.6 The timing-effect frontier

Figure 9 makes the timing point quantitatively explicit, illustratively. For interventions of equal nominal content, the effect size as a function of offset from a state-transition window has a sharp peak. The state-aware delivery sits near the peak. The correlational baseline, which delivers interventions on engagement triggers, sits in the low-effect regime almost everywhere. The ratio between optimal-timing and mistimed delivery is in the range that the just-in-time adaptive intervention literature has documented for behaviour change [25]. The implication for system design is that timing is not a small refinement on top of content optimisation. It is the larger lever.

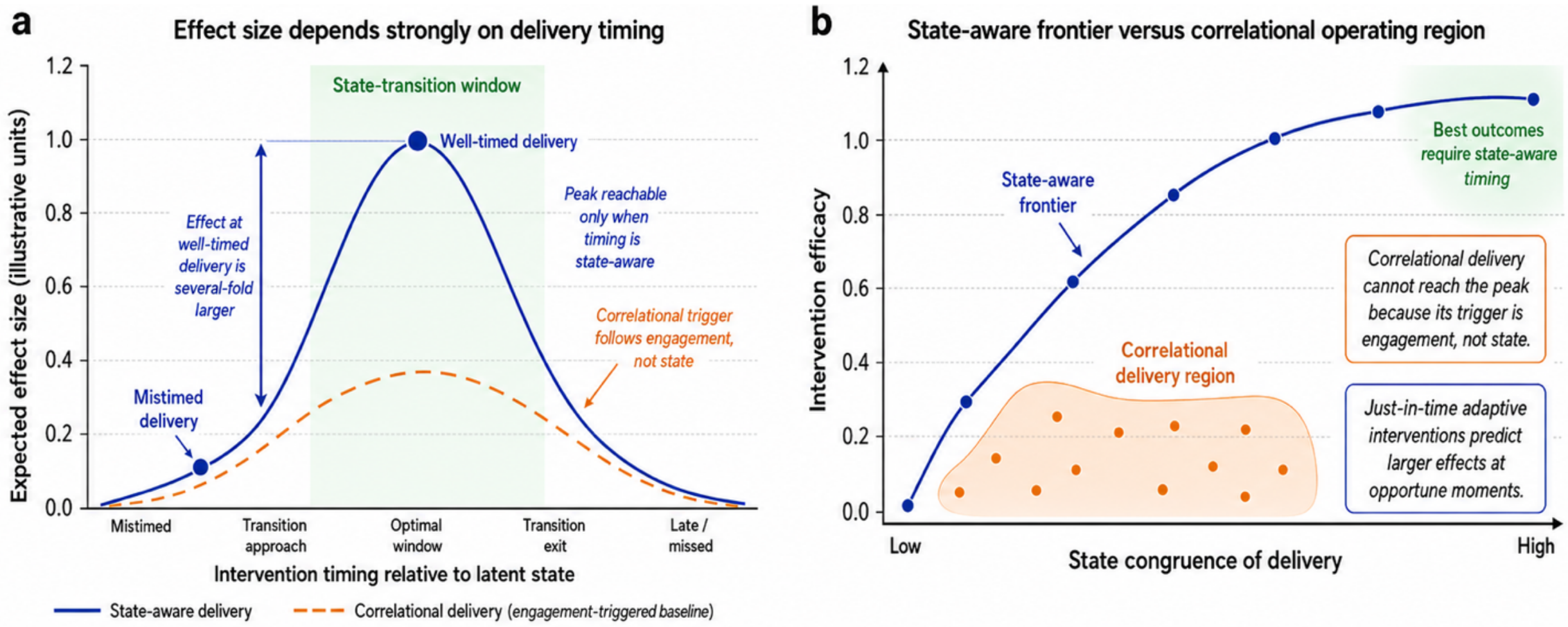

**Figure 9.** Intervention timing and effect size: state-aware versus correlational delivery. The frontier is illustrative and qualitatively consistent with the just-in-time adaptive intervention literature [25]. The effect size at well-timed delivery is several times larger than at mistimed delivery. The correlational baseline cannot reach the peak because its trigger is engagement, not state.

## 5.7 State as a Markov structure

A useful coarse-grained model of state dynamics is a Markov chain over a small set of coarse state regions. Figure 10 illustrates the structure. Some regions are stable: arousal-low rest, recovery, active engagement. Others are transition regions: rising-arousal, recovery-to-action. The transition regions are precisely the windows in which intervention has high leverage. The same dynamic that makes these regions transient also makes them targets. A state-aware system identifies them and acts in them. A correlational baseline does not see them as different from any other moment of high engagement.

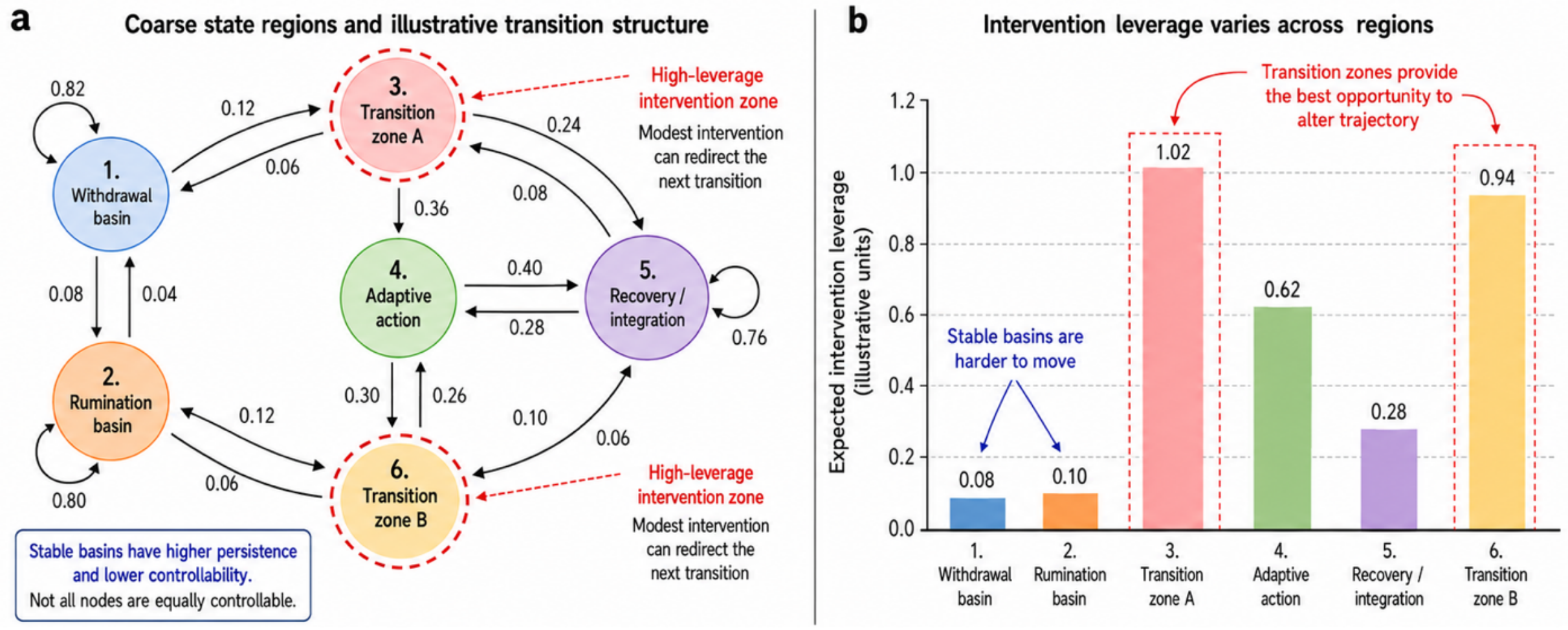

**Figure 10.** *Simplified state-transition Markov structure (illustrative). Nodes are coarse state regions. Edge weights are illustrative transition probabilities. Red dashed rings mark transition zones in which intervention has high leverage. Real state dynamics are higher-dimensional and partly continuous, but the Markov picture captures the structural point: not all regions are equally controllable.*

## 5.8 Three qualitative patterns from the observational base

Without claiming any of these as quantitative findings, three qualitative patterns recur in the deployment data and merit description. We note them because they are the patterns that informed the framework, and because they suggest the empirical structure that the companion paper will test.

***Pattern one: within-week regularities are stronger than across-week regularities.*** When a user's outcome trace is examined across a week, day-of-week effects emerge that are stable across many weeks. Mondays look like Mondays. Friday evenings look like Friday evenings. The structural correlate is that the state-trajectory has a weekly envelope that constrains how a generic event will land on a given day, before any individual feature of the day is considered. Personalisation systems that ignore the day-of-week structure are predicting against this envelope without using it.

***Pattern two: outcomes after intervention show a bimodal distribution under timing variation.*** When intervention exposure is examined conditional on its timing relative to the user's typical state-transition windows, the outcome distribution is not unimodal. Well-timed exposures cluster in a high-effect mode. Mistimed exposures cluster in a low- or null-effect mode. The bimodality is what one would expect if Figure 9 captures the underlying frontier even approximately. Engagement-triggered systems that average across the timing dimension are averaging over the bimodal distribution and losing most of the signal.

***Pattern three: persona-level differences are smaller than within-persona differences.*** Across the four personas, the modal outcome class for a given event differs in the direction Figure 7 illustrates, but the within-persona variance is consistently the larger source of outcome variation. This is consistent with the central claim of the paper: that state, not persona, is the variable that does the heavy lifting. Personalisation systems that segment by persona and then apply persona-averaged interventions are leaving the larger source of variance on the table.

## 6 Six Operational Requirements for a State-Aware System

The framework above implies that any deployed system that intends to support state-conditional intervention must operationally support six things. These are not architectural choices. They are properties any architecture must instantiate to be a candidate. We list them and contrast with what a correlational baseline would and would not support.

State-causation pipeline
biology
physiology
neuropsychology
State $S_t$
Decision / action $A_t$
Outcome $Y_{t+1}$
Updated state $S_{t+1}$

| Capability | State-aware system | Correlational baseline | Operational signal |
|---|---|---|---|
| C1 Real-time state estimation | individual-indexed, updated continuously | population averages, refreshed periodically | within-day prediction calibration |
| C2 Causal-path attribution | identifies which channel drove the outcome | reports correlated features, no attribution | intervention-specific effect recovery |
| C3 Intervention timing | targets state transition windows | fires on engagement triggers | effect size as a function of timing |
| C4 Outcome forecasting | per-individual trajectory under do(a) | group-level base-rate prediction | individual hit-rate over horizon |
| C5 Counterfactual replay | reconstructs what would have happened | unavailable | auditability score |
| C6 Welfare separation | distal goal distinct from proximal engagement | objective collapses to engagement | goal-engagement decoupling index |

**Figure 11.** *What a state-aware system must operationally support. Six capability requirements (C1 to C6) follow from the state-causation framework. The right-most column suggests measurable signals by which a system can be audited against the requirements.*

**C1. Real-time state estimation.** The system must maintain an individual-indexed estimate of the current state at a cadence that matches the cadence at which states actually change. For a behavioural-platform deployment, this is closer to hourly than daily. A correlational baseline that updates user features on a session-by-session or week-by-week cadence is, by construction, out of phase with the dynamics it is trying to predict.

**C2. Causal-path attribution.** When an outcome diverges from what was expected, the system must be able to attribute the divergence to a channel: biological, physiological, neuropsychological, or contextual. Without causal-path attribution, the system can identify what changed but not why. This is the gap that correlational personalisation engines have made famous and that doomed the cohort of recommender-based wellbeing applications of the early 2020s.

**C3.** Intervention timing. The system must target intervention to state-transition windows, not to engagement triggers. This is a behaviour the system must learn from longitudinal data on the individual, not a parameter that can be set globally. Effect sizes for the same intervention delivered at well-timed versus mistimed moments differ by an order of magnitude or more in the literature on behaviour change [25, 32].

**C4. Per-individual outcome forecasting.** The system must produce probabilistic forecasts of the next-period outcome distribution under specified interventions. Forecasts must be at the individual level, not the cohort level. Cohort-level forecasts are a category error here because they collapse the variance the framework is trying to capture. Figure 12 shows what the predicted calibration profile looks like for the two architectures.

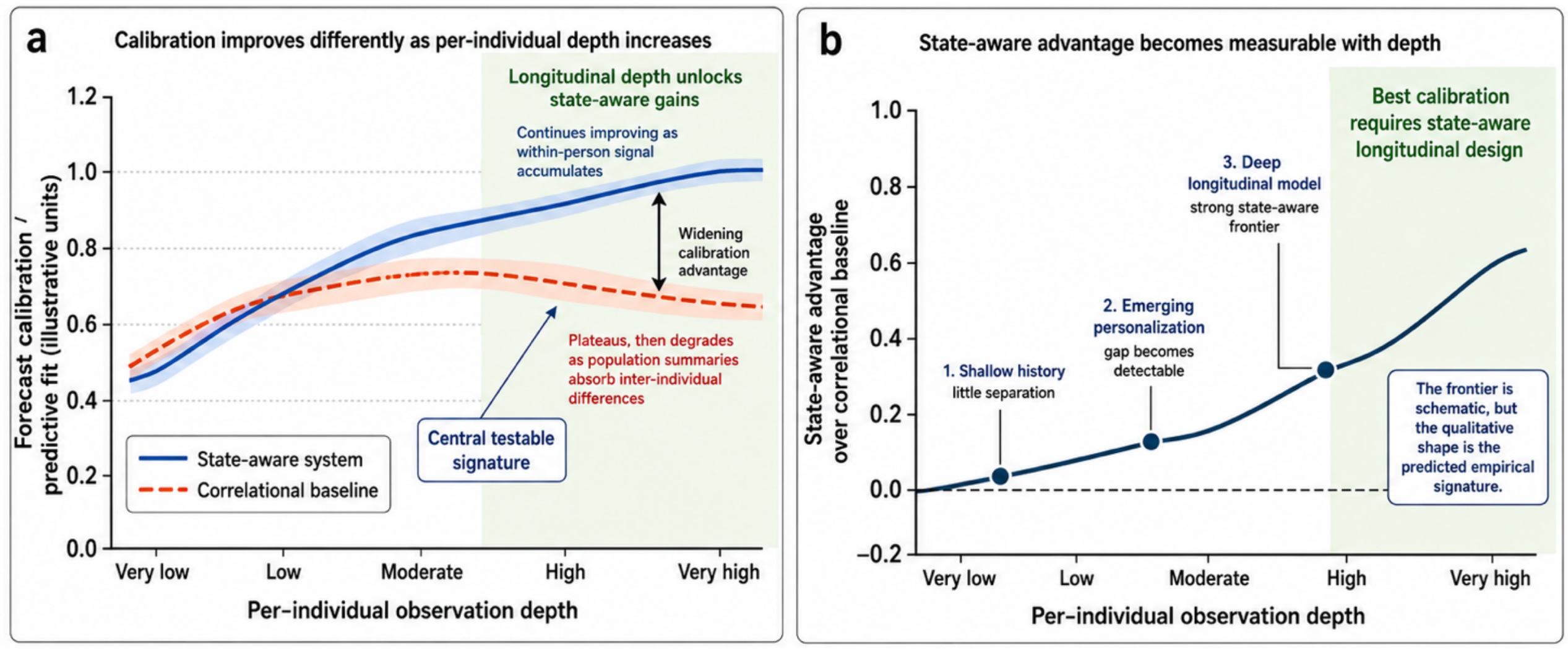

**Figure 12.** *Forecast calibration over increasing per-individual observation depth. State-aware systems continue to improve as longitudinal data accumulates. Correlational baselines plateau and then degrade as inter-individual differences are absorbed into population estimates. The schematic frontier is the central testable signature of state-aware design.*

**C5. Counterfactual replay.** The system must be able to reconstruct, for a past episode, what would have happened under an alternative intervention. This is the only operational signature of genuine causal inference at the individual level. Everything else is consistent with high-quality association. The counterfactual must be evaluable against subsequent observation when state-similar episodes recur.

**C6. Welfare separation.** The system's objective must keep proximal engagement signals (clicks, dwell, sentiment of self-report) structurally separate from distal welfare signals (long-horizon goal attainment, stability of trajectory, intervention-dependent gains). A system that collapses welfare into engagement cannot, by construction, be aligned with the user's long-horizon state trajectory.

Each of these six requirements is in principle satisfiable. The hard problems are not the requirements themselves. They are the data infrastructure, the architectural choices that propagate state through the channels, the latency budget for real-time estimation, and the ethical constraints on welfare elicitation. Section 9 returns to these as the agenda for future work.

## 7 Seven Testable Predictions

The framework is committed to a research programme rather than a single hypothesis. We commit it to seven specific, falsifiable predictions that distinguish state-aware from correlational accounts of human outcomes. Each is followed by an operationalisation.

**Prediction P1 (State-conditional outcome divergence).** Within-person variance in outcome class for a fixed event class is larger than between-person variance for the same event class, after controlling for stable individual covariates. State-aware models that include time-indexed weighting estimates will capture this within-person variance. Correlational baselines will not.

*Test. Variance decomposition on longitudinal outcome data using nested random-effects models.*

**Prediction P2 (Channel attribution recovery).** When state, biology, physiology, neuropsychology, and outcome are jointly observed, a state-aware model can recover the channel through which a given outcome divergence propagated. A correlational model can recover associations but not the directed path.

*Test. Mediation analysis with channel-specific instruments. Recovery of known structural relations on synthetic data with ground truth.*

**Prediction P3 (Intervention timing dominates intervention content).** For interventions of equal nominal content, effect size as a function of state-transition proximity is larger than effect size as a function of content variant. State-aware timing dominates content optimisation.

*Test. Factorial micro-randomised trials varying timing and content independently.*

**Prediction P4 (Per-individual forecast calibration).** A state-aware system produces individual-level outcome forecasts whose calibration improves as observation horizon lengthens. A correlational baseline's calibration plateaus quickly and then degrades as inter-individual differences are absorbed into population estimates.

*Test. Longitudinal calibration curves on held-out individuals at increasing observation depths.*

**Prediction P5 (Counterfactual coherence).** For sequences in which an individual experienced two state-similar episodes with different interventions, a state-aware system can predict the second outcome from the first episode's state and counterfactual replay. The reverse direction (predicting the first from the second) is also stable. Correlational baselines fail this symmetry test.

*Test. Within-individual paired-episode prediction with intervention swap.*

**Prediction P6 (Welfare-engagement decoupling).** State-aware systems can be configured to optimise for distal welfare signals at controlled cost to proximal engagement. Correlational personalisation systems face a steep frontier on the same trade-off because their objective is structurally entangled.

*Test. Pareto-frontier comparison on multi-objective deployments under elicited welfare targets.*

**Prediction P7 (Cross-persona transport).** The within-individual state dynamics generalise across personas in their structural form (the dimensions, the channels, the transition behaviour) even though the parameter values do not. Correlational models trained on one persona transfer poorly to another. State-aware models with shared structure and persona-specific parameters transfer well.

*Test. Held-out persona transfer with shared structural prior and persona-conditional fine-tuning.*

These predictions are inter-related but logically independent. Each can in principle be falsified without the others. We invite the field to construct shared longitudinal benchmarks that test them.

## 8 Implications and Limitations

### 8.1 For digital health

Digital health applications, particularly in mental health, are dominated by content-personalisation engines built on engagement objectives. The framework here implies that those engines cannot, by construction, deliver welfare-aligned outcomes at scale. Their objective is in the wrong layer of the causal chain, and their personalisation operates on stable user features rather than on time-indexed state. A state-aware redesign is feasible but requires a different data infrastructure (per-individual longitudinal state estimation), a different objective (welfare with structural separation from engagement), and a different intervention model (timing-first, not content-first). The implications for clinical decision-support are parallel. A state-aware decision-support layer would not replace clinical judgement, but it would surface state-transition windows in which a brief evidence-based intervention is likely to have outsized effect.

### 8.2 For education

The same logic applies in educational personalisation. The dominant model treats the learner as a knowledge-state vector and personalises content accordingly. The framework here implies that the relevant state is broader (including attentional and affective dimensions) and faster-moving (within-session rather than across-session). The right intervention is often not a content swap but a state-respecting decision about whether to introduce the next piece of content at all. The empirical evidence for state-conditional learning is well-established in the spacing-effect and desirable-difficulty literature [3, 33]. The gap is in personalisation engines that act on it.

### 8.3 For AI personalisation

The broad implication for the AI personalisation industry is structural rather than incremental. The default architecture (user-features-times-context predicts engagement) is in the wrong layer for the problem it claims to solve. Adding causal-inference modules on top of a fundamentally associational pipeline is a category mismatch. The right move is to put state at the centre and let association become a special case of state-conditional inference. Biswas [2] developed the system-specification version of this argument. The present paper develops the person-side argument that motivates it.

### 8.4 For personal agency

The framework supports a meaningful sense in which a person can control their outcomes. Not by willing the outcome. The decision-layer is the weakest lever. Not by adding effort. By tracking their own state, recognising state-transition windows, and intervening on the upstream channels where leverage is highest. This is not a new idea in the contemplative and clinical traditions. It is rarely formalised in a way that connects to digital tools. Four practical implications follow. Track the state at a cadence finer than the day. Respect the bandwidth gap by not over-relying on verbal narrative for self-attribution. Choose interventions on the biological and physiological channels for fast-acting leverage and on the neuropsychological channel for slow-acting structural change. Choose intervention timing as carefully as intervention content.

### 8.5 Connection to behavioural economics

A reader from behavioural economics will recognise the family resemblance to dual-process accounts of decision-making [15]. The state-weighting framework is consistent with the dual-process taxonomy but goes beyond it in two ways. First, the System 1 / System 2 distinction in behavioural economics has been criticised for being too binary and for treating system one as a residual category for everything that is not deliberation [29]. The state-weighting framework replaces the binary distinction with a continuous weighting variable that takes different values for different events, in different states, at different times. Second, the prescriptive component of behavioural economics (nudge architecture) operates by changing the choice environment around a decision. The framework here suggests that the choice environment is one of several channels through which a state can be moved. Environment-design is a real lever, but it is a co-equal partner with biological, physiological, and neuropsychological intervention rather than the privileged one.

### 8.6 Limitations

We list the limitations explicitly. First, the framework is a position contribution, not an empirical finding. The observational base is real, but the quantitative analysis is deferred. Second, the cohort is heavily skewed toward students, and the cross-persona generalisations should be treated as conjectures pending replication. Third, the platform does not collect clinical measurements. Biological and physiological claims in the framework are theoretically grounded but not validated against gold-standard instruments in our own data. Fourth, the framework is implementation-neutral. It specifies what is required of a state-aware system but not how to build one. The choices that turn the framework into a deployable artefact are non-trivial and are the subject of separate technical work.

Fifth, the conditional sense of control we defend is bounded. A person whose state is in a low-controllability window cannot will themselves into a state-transition. The framework supports moments of greater agency, not constant agency.

### 8.7 Ethical considerations

Two ethical observations follow. First, a state-aware system is by construction a more powerful instrument than a correlational one. It identifies windows in which the user is most amenable to behavioural change. The same instrument that supports welfare can in principle be used to extract engagement. The welfare-separation requirement (C6) is therefore an ethical requirement, not merely an engineering one. Second, the framework implies that users have an interest in the structural identifiability of the systems that model them. A user whose state is being inferred has a meaningful claim to know which channel a recommendation is acting on, with what evidence, and against what objective. The architectural opacity of contemporary personalisation systems is not an implementation detail. It is a barrier to contestation.

## 9 Future Work

We treat this paper as a position contribution at the start of a multi-year research programme. The agenda has four parts.

### 9.1 Empirical companion

The first piece of follow-on work is the empirical companion paper. The plan is to draw from the 24-month observational base described in Section 5, define a pre-registered analysis plan on a held-out portion of the data, and report quantitative findings on a small set of the seven predictions in Section 7. The disclosure boundary is the constraint. We will report effect sizes, confidence intervals, and per-persona variance decomposition. We will not report architectural details that fall within the scope of the patent applications.

### 9.2 Cross-deployment replication

A single-deployment observational base cannot support the strongest claims the framework makes. The second piece of follow-on work is replication across deployments that share the structural commitments (P1 to P4 in Biswas [2]) but differ in cohort, intervention library, and content domain. We are in early conversation with partner institutions on three such replications, one in education, one in occupational health, and one in adolescent wellbeing.

### 9.3 Validated instruments

The platform does not currently collect data from validated neuropsychological instruments. The third piece of follow-on work is to integrate brief validated instruments into the consent flow so that platform-derived state estimates can be benchmarked against established measures. Candidate instruments include short forms of the PHQ-9, GAD-7, and PSS for the affective channel, and brief cognitive tasks for the attentional channel. The integration has to respect the user experience and the consent regime. Both are non-trivial.

### 9.4 Methodological work

The fourth piece is methodological. The framework places identifiability demands on the modelling layer that are not satisfied by off-the-shelf machine-learning architectures. We are developing the necessary structural commitments in parallel with empirical work. That methodological work is the subject of separate publications and patent applications and is not described here.

## 10 Conclusion

The puzzle we began with was that the same person, given the same input, does different things on different occasions. The position we have defended is that the variability is the signature of a real and characterisable feature of human cognition. The latent state, regarded as a time-indexed weighting vector over biological, physiological, and neuropsychological dimensions, determines how the next event is processed and what outcome it produces. The relationship is causal, not correlational. The conscious channel through which decisions are reported is narrow, biased, and a poor place to intervene. The leverage on outcomes lies upstream of the verbal narrative, in the channels that propagate the state forward to the decision. A person can in this sense control their outcomes, conditional on their state-trajectory and the timing of intervention. A system that supports this kind of control must operationally satisfy six requirements that contemporary correlational personalisation engines do not satisfy.

The framework is a position, not an empirical study. It is grounded in established literature across six fields and motivated by a 24-month deployment context spanning research from 2023 to 2026. The seven predictions in Section 7 are intended to commit the framework to falsifiable empirical work. The operational requirements in Section 6 are intended to commit it to deployment-relevant engineering. Both commitments are open, and we invite scrutiny.

## Author contributions

S.B. conceived the framework, developed the formal definitions and the four-channel causal model, and led the writing. S.G. led the design of the observational base, the data infrastructure that supports the platform, and the analytical pipeline that produced the qualitative patterns described in Section 5. P.M. contributed as an AI Researcher at Dots-In, supporting AI research synthesis, manuscript structuring, figure and table review, citation-numbering checks, and review of the operational system requirements. All authors contributed to the operational requirements in Section 6 and to the predictions in Section 7. All authors approved the final manuscript.

## Competing interests

S.B. and S.G. are co-founders and equity holders in the entity that operates the behavioural platform from which the observational base in Section 5 is drawn. P.M. is an AI Researcher at Dots-In. The methodological commitments described in the framework are the subject of separate patent applications. No external funding was received for the writing of this paper.

## AI use disclosure

Drafting of this manuscript was assisted by a large language model. All scientific claims, framings, references, and arguments are the authors' own and have been independently verified. The framework and the empirical observations reported are the authors' own. The language model was used as a writing and structuring aid, not as a source of evidence or argument.